\documentclass[conference, letterpaper]{IEEEtran}
\usepackage[top=0.72in, bottom=1in, left=0.62in, right=0.62in]{geometry}

\IEEEoverridecommandlockouts
\usepackage{amsmath}
\usepackage{graphicx}
\usepackage{cite}
\usepackage{amsmath,amssymb,amsfonts, amsthm}
\usepackage[linesnumbered,ruled]{algorithm2e}
\usepackage{algorithmicx}
\usepackage{algpseudocode}
\usepackage{graphicx}
\usepackage{textcomp}
\usepackage{xcolor}
\usepackage{xspace}
\usepackage{soul}
\usepackage{kotex}

\usepackage{tcolorbox}
\usepackage{float}  
\usepackage{caption} 

\usepackage{booktabs} 
\usepackage[normalem]{ulem} 
\usepackage{ragged2e}

\usepackage{xspace}
\newcommand{\BfPara}[1]{{\noindent\bf#1.}\xspace}
\usepackage{tabularx}
\usepackage{color}
\definecolor{pgreen}{rgb}{0,0.5,0}
\definecolor{pred}{rgb}{0.9,0,0}
\definecolor{ppurple}{rgb}{0.5,0,0.5}

\usepackage{subfigure}
\usepackage{xcolor}
\usepackage{tikz}
\usepackage{listings}
\usepackage{verbatim}

\usepackage[T1]{fontenc}
\usepackage[utf8]{inputenc}

\definecolor{color1}{RGB}{228,26,28}
\definecolor{color2}{RGB}{55,126,184}
\definecolor{color3}{RGB}{77,175,74}
\definecolor{color4}{RGB}{152,78,163}

\usepackage{float}

\definecolor{promptbg}{HTML}{F2F2FF}      
\definecolor{promptborder}{HTML}{808080}  

\usepackage{amsmath,amsfonts,amssymb}
\usepackage{tabularx}
\usepackage{tikz}

\begin{document}
\title{Quantum Circuit Structure Optimization for Quantum Reinforcement Learning}

\author{\IEEEauthorblockN{
    Seok Bin Son and 
    Joongheon Kim
    \thanks{This research was supported by the MSIT (Ministry of Science and ICT), Korea, under the ITRC (Information Technology Research Center) support program (IITP-2024-RS-2024-00436887) supervised by the IITP (Institute for Information \& Communications Technology Planning \& Evaluation); and also by IITP grant funded by MSIT (RS-2024-00439803, SW Star Lab). \textit{(Corresponding author: Joongheon Kim)}}
}
\IEEEauthorblockA{
Department of Electrical and Computer Engineering, Korea University, Seoul, Republic of Korea \\
E-mails: 
\texttt{\{lydiasb,joongheon\}@korea.ac.kr} 
}
}

\maketitle
\begin{abstract}

Reinforcement learning (RL) enables agents to learn optimal policies through environmental interaction. However, RL suffers from reduced learning efficiency due to the curse of dimensionality in high-dimensional spaces. Quantum reinforcement learning (QRL) addresses this issue by leveraging superposition and entanglement in quantum computing, allowing efficient handling of high-dimensional problems with fewer resources. QRL combines quantum neural networks (QNNs) with RL, where the parameterized quantum circuit (PQC) acts as the core computational module. The PQC performs linear and nonlinear transformations through gate operations, similar to hidden layers in classical neural networks. Previous QRL studies, however, have used fixed PQC structures based on empirical intuition without verifying their optimality. This paper proposes a QRL-NAS algorithm that integrates quantum neural architecture search (QNAS) to optimize PQC structures within QRL. Experiments demonstrate that QRL-NAS achieves higher rewards than QRL with fixed circuits, validating its effectiveness and practical utility.
\end{abstract}

\section{Introduction}\label{sec:intro}


Reinforcement learning (RL) has achieved remarkable progress across various application domains based on classical neural networks (NN). NN-based RL has been successfully applied to game playing, robot control, autonomous driving, and satellite communication systems~\cite{10506797, 10994421, 9364754}. It often surpasses human-level performance even in complex environments. However, existing RL faces fundamental limitations when handling high-dimensional state and action spaces~\cite{10232949}. As state and action dimensions increase, the number of learnable parameters grows exponentially, resulting in the curse of dimensionality~\cite{10889145}. This exponential growth significantly reduces learning convergence speed and computational efficiency. Moreover, data sparsity in high-dimensional spaces requires massive sample sizes to learn optimal policies~\cite{10889145}, imposing substantial temporal and cost constraints on real-world systems.

\begin{figure}[t!]
    \begin{center}
        \includegraphics[width=3.5in]{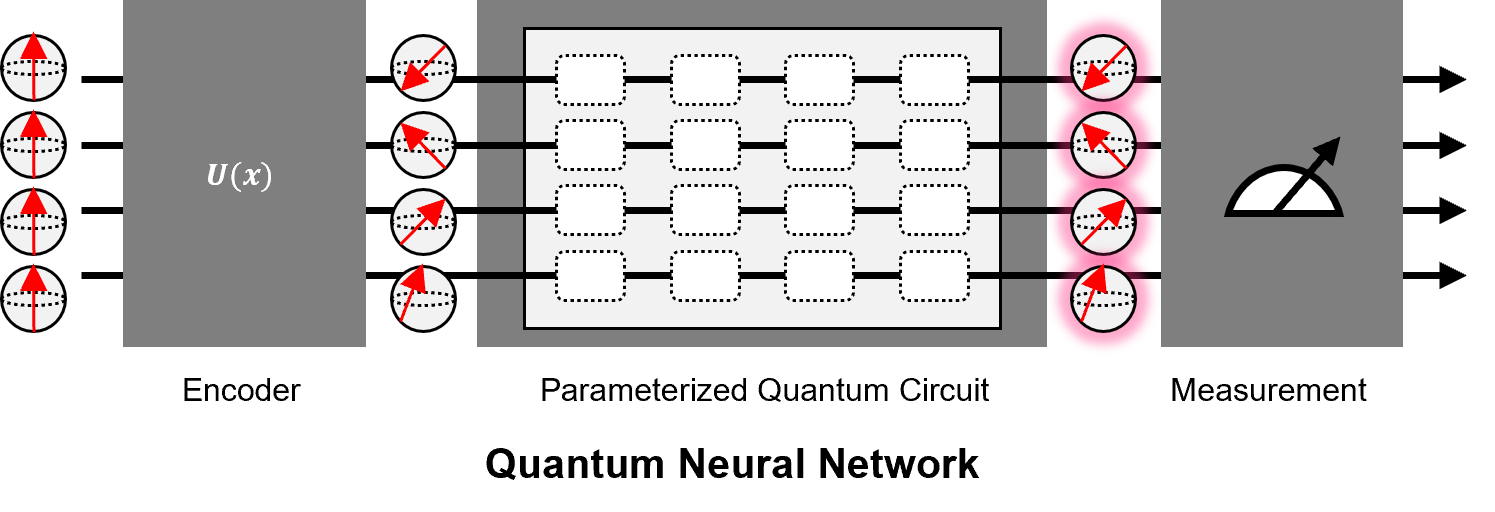}
    \end{center}
    \caption{The structure of quantum neural network.}
    \label{fig:QNN}
    \vspace{-3mm}
\end{figure}

To overcome these structural limitations, quantum reinforcement learning (QRL) has gained increasing attention~\cite{QRL_1, QRL_2, QRL_3}. QRL utilizes quantum neural networks (QNNs) to exploit quantum computing properties such as superposition and entanglement~\cite{QRL_1}. As illustrated in Fig.\ref{fig:QNN}, QNNs consist of three components: encoder, parameterized quantum circuit (PQC), and measurement\cite{10738386}. The encoder converts classical inputs into quantum states for processing within the quantum circuit. The PQC performs linear and nonlinear transformations similar to hidden layers in classical NNs. The measurement stage converts quantum states into classical outputs for verification. By using QNNs, QRL addresses the curse of dimensionality and sample inefficiency inherent in NN-based RL. For example, QNNs can compressively represent high-dimensional data using a small number of qubits, reducing computational costs. Furthermore, entanglement enables QNNs to achieve high performance with fewer training samples~\cite{QRL_3}. Due to these advantages, QRL is emerging as a powerful alternative in complex environments such as high-dimensional satellite communication networks, multi-UAV cooperative systems, and smart factories~\cite{QRL_1, QRL_2, QRL_3}.

However, important limitations remain unresolved in current QRL research. One critical issue concerns the optimality of the PQC structure, which plays a key role in determining QRL performance. Most existing QRL studies have relied on researchers' empirical intuition or conventional circuit patterns when designing PQCs. Typically, combinations of a few $RX, RY$, and $RZ$ rotation gates with entanglement gates such as $CX$ are adopted. Repeatedly stacking single-type PQC blocks is also a common practice. Nevertheless, such passive and fixed designs do not reflect the complexity and variability of real-world environments. This design limitation inherently restricts the ability to ensure optimal circuit structures for specific problems. The PQC structure directly influences learning stability, convergence speed, and the final policy performance of agents. Furthermore, gate arrangement and parameter configuration determine the expressiveness and function approximation capability of quantum circuits. Inefficient structures also increase the accumulation of quantum noise. As a result, suboptimal PQC designs degrade learning performance and sharply raise computational costs.


To maximize the potential of QRL, recent studies have recognized the need for automatic PQC optimization beyond empirical design. Specifically, it is necessary to automatically search and select gate types, placements, and depths based on environmental characteristics and learning objectives. This enables the design of optimal circuits that balance expressiveness and computational efficiency. This paper proposes a methodology that introduces neural architecture search (NAS) into PQC design for QRL to address these limitations. NAS is a technique widely used in deep learning to automatically explore and optimize model structures based on data~\cite{NAS_1, NAS_2, NAS_3}. It replaces manual design reliant on researcher intuition and significantly improves model performance and efficiency. The core idea is to apply NAS's exploration capabilities to quantum circuit design problems. This enables interactive exploration of gate combinations and layout structures within PQCs in real time in RL environments. Ultimately, this approach derives optimal circuits that minimize unnecessary or redundant gate usage while maximizing computational efficiency and policy performance. By adopting this methodology, suboptimal PQC design issues in existing QRL can be addressed. This allows agents to achieve higher cumulative rewards and enhanced learning performance.

The main contributions of this paper are as follow,

\begin{itemize}

    \item This paper proposes an approach that integrates NAS into QRL to overcome the limitations of existing empirically designed QRL methods. By automatically exploring and optimizing PQC structures, this approach presents a new research direction for designing quantum circuits optimized for specific environments.
    

    \item A comprehensive set of gate candidates was constructed, including single-qubit gates (i.e., U3, RX, RY, RZ) and two-qubit gates (i.e., CU3, SWAP, CX, CY, CZ), to maximize the expressiveness and versatility of NAS exploration. This enabled efficient exploration of diverse quantum gate combinations and layout structures.


    \item Comparison of the proposed QRL-NAS framework with existing QRL using fixed PQC structures showed superior performance in policy cumulative rewards. These results empirically demonstrate the importance of PQC structure optimization in enhancing QRL performance and validate NAS as a practically effective approach in QRL design.

\end{itemize}

\section{Preliminaries}\label{sec_2:Preliminaries}

\subsection{Neural Architecture Search} 

\begin{figure}[t!]
    \begin{center}
        \includegraphics[width=3.5in]{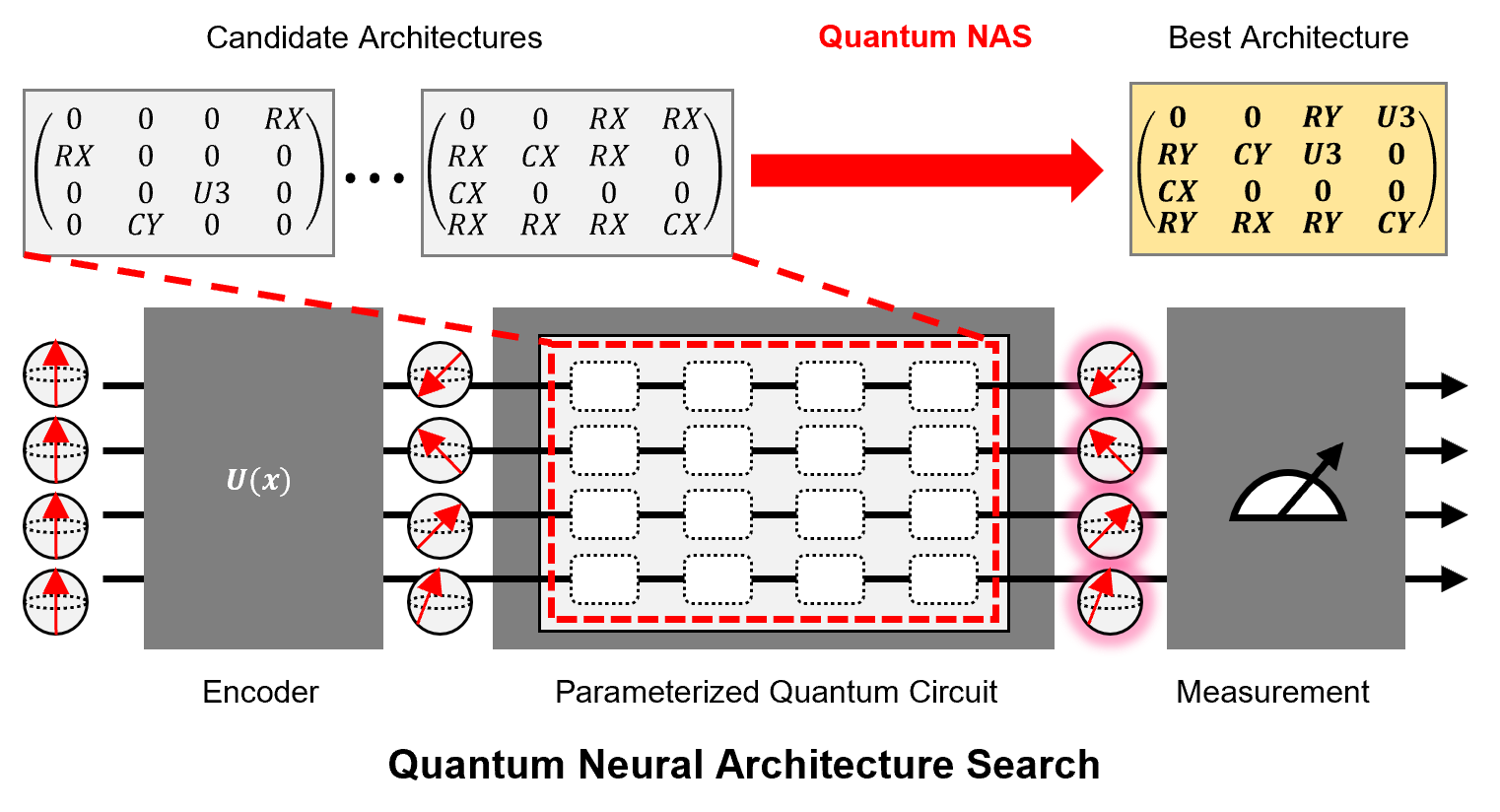}
    \end{center}
    \caption{The structure of quantum neural architecture search.}
    \label{fig:QNAS}
    \vspace{-3mm}
\end{figure}

Traditionally, NN structures were manually designed by machine learning experts and repeatedly tested to verify their performance. However, this manual approach was inefficient due to substantial time and computational costs. NAS was developed to address this inefficiency by automating the design process. NAS automatically searches for NN architectures with desired performance, establishing itself as a core AutoML technology~\cite{DBLP:conf/ictc/KimYLJK21}. A NAS framework generally comprises two core components: (i) the search space defining the set of candidate architectures to explore, and (ii) the search strategy responsible for selecting, training, and evaluating these architectures~\cite{DBLP:conf/ictc/KimYLJK21}. The performance and efficiency of NAS depend heavily on the design of these two components, as the search space determines the scope of possible architectures, while the search strategy governs exploration efficiency and optimization accuracy. Recent NAS research has explored various design paradigms, including cell-based search spaces, hierarchical search spaces, and differentiable search strategies, to improve scalability and computational feasibility. In this paper, the exploration capabilities of NAS are applied to QNNs, as illustrated in Fig.~\ref{fig:QNAS}. This enables automatic searching for quantum circuit structures optimized for specific problem environments in QRL. By integrating NAS with QRL, this approach enables the automatic design of PQC structures that maximize both expressiveness and computational efficiency. This represents a novel research direction that overcomes the limitations of traditional empirically designed quantum circuits, potentially enhancing learning stability, convergence speed, and policy performance in complex environments.

\begin{figure*}[t!]
\centering
\includegraphics[width=1\textwidth]{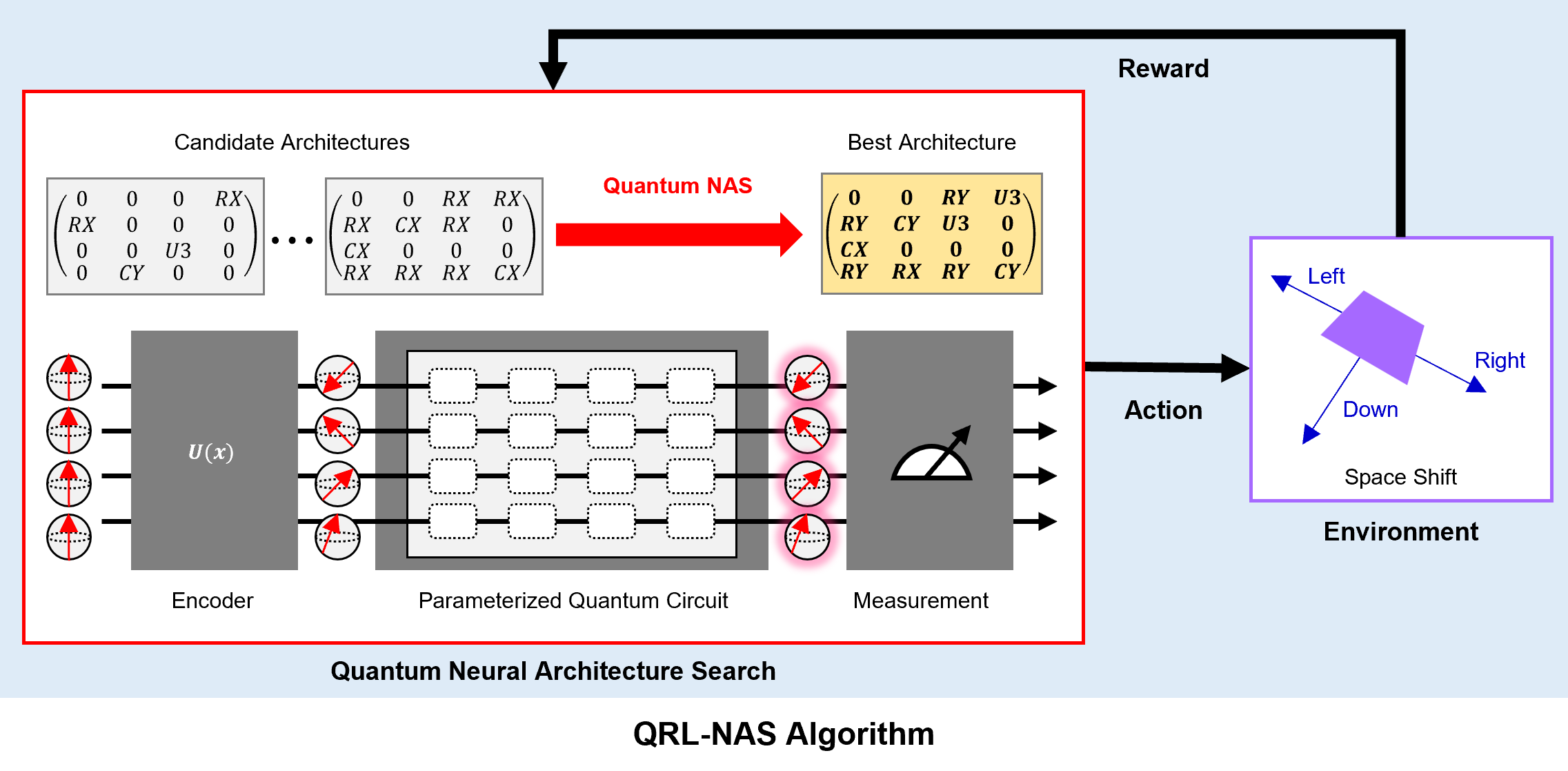}
\caption{The overall architecture of QRL-NAS.}\label{fig:QRL_NAS}
\end{figure*}

\subsection{Reinforcement Learning} 

RL is a core technique widely used to solve complex decision-making problems by learning optimal action strategies in various environments. This approach improves policies to maximize rewards through interactions between agents and their environments. Representative RL techniques include the value-based deep Q-Network (DQN)~\cite{DQN} and the policy-based reinforce algorithm~\cite{williams1992simple}. DQN combines Q-learning with deep learning to approximate state-action value functions in high-dimensional state spaces. This allows agents to calculate Q-values for each state and apply an $\epsilon$-greedy policy to balance exploration and exploitation. However, DQN suffers from the curse of dimensionality, as learning becomes exponentially difficult with increasing state and action space sizes. Reinforce, a policy gradient method, updates parameters by multiplying the log probability of chosen actions by the received rewards. This enables direct optimization of policies without Q-value estimation, making it suitable for continuous or high-dimensional action spaces. However, reinforce often exhibits reduced learning stability due to high variance in gradient estimation. Recent research has focused on integrating RL techniques with quantum circuit-based QRL to enhance performance. Nonetheless, most existing QRL methods employ fixed quantum circuit structures, limiting their adaptability to specific problem environments. To address these limitations, this paper proposes a QRL-NAS algorithm that integrates NAS to optimize quantum circuit structures for QRL.



\section{Proposed Algorithm: QRL-NAS} \label{sec:Algorithm}

\subsection{Design Principles}

The QRL-NAS algorithm proposed in this paper integrates NAS techniques to automatically optimize quantum circuit structures, maximizing the performance and efficiency of QRL. Previous QRL studies have primarily designed PQC structures based on researchers' empirical intuition and fixed circuit templates. For instance, single-qubit rotation gates such as $RX, RY$, and $RZ$, along with entanglement gates like $CX$, were typically arranged in predetermined orders. In many cases, identical PQC blocks were repeatedly configured across layers. However, this fixed circuit design approach presents several limitations. Structures that do not reflect the characteristics of the problem environment fail to sufficiently model the complexity of target policies, limiting the expressiveness of state-action mappings. Moreover, using the same circuit configuration across environments often includes unnecessary gates, increasing computational costs and reducing execution efficiency. Excessively restricted or complex circuit structures may also slow convergence speed or accumulate quantum noise, significantly degrading policy performance. Therefore, the fixed circuit design of the existing QRL acts as a fundamental constraint that impairs expressiveness and computational efficiency. It also limits the generalization capability of learned policies by failing to explore circuit structures optimized for the problem environment.


This paper proposes QRL-NAS, a methodology that improves QRL performance and efficiency by integrating NAS to overcome existing limitations. QRL-NAS automatically explores and optimizes the gate types, placement order, and depth of PQC while interacting with the environment to design problem-optimized circuit structures. To achieve this, QRL-NAS constructs a comprehensive candidate space comprising single-qubit gates (i.e., U3, RX, RY, RZ) and two-qubit gates (i.e., CU3, SWAP, CNOT, CY, CZ). The operation flow of QRL-NAS is as follows. The agent observes environmental states and embeds them into quantum states using angle encoding in the encoder module. NAS is then performed within PQC, sampling candidate gates at each location to select suitable gate combinations and arrangements for the current situation. The PQC designed in this manner operates as the agent's policy network, outputting actions through measurement. Selected actions are applied to the environment to generate new states and rewards. Rewards and state transitions obtained in this process are used for Q-value updates based on DQN and for NAS search feedback. This enables iterative optimization of both the PQC structure and its parameters.

\subsection{Framework of QRL-NAS}

\BfPara{State Space} In this algorithm, the state is defined as the observations obtained by the RL agent from the environment. In the LunarLander-v2 environment, the state consists of eight elements: x and y coordinates of the lander, horizontal and vertical velocities, angle, angular velocity, and contact status of the left and right legs with the ground. This state information is embedded into the quantum circuit input via the encoder using an $RX$ gate-based angle encoding method. The resulting quantum representation enables the PQC to predict optimal actions.

\BfPara{Action Space} The action of QRL-NAS is defined by selecting gates to apply to each qubit within the quantum circuit structure. Gate selection includes single-qubit gates (i.e., U3, RX, RY, RZ) and two-qubit gates (i.e., CU3, SWAP, CNOT, CY, CZ), determined through NAS exploration. After processing by the PQC, measurement results are interpreted as Q-values for each action. Final actions are then selected using an $\epsilon$-greedy policy.

\BfPara{Reward Design} Rewards are defined as values obtained from the outcomes of an agent's actions in the environment. In the LunarLander-v2 environment, the lander receives high rewards for safe and accurate landings and negative rewards for crashes or unstable landings. QRL-NAS optimizes the quantum circuit structure to maximize these rewards. The algorithm is designed to select gate combinations that improve both learning convergence and the expected policy reward.

\section{Performance Evaluation}

\subsection{Simulation Settings}

This paper verifies the performance of the proposed QRL-NAS by comparing it with existing QRL algorithms. The LunarLander-v2 environment, a representative RL benchmark, was used as the experimental environment for all algorithms. All quantum circuit implementations were conducted using the \textit{TorchQuantum} library with a four-qubit setup. In the experiments, the learning rate was set to 0.1, the discount factor $\gamma$ to 0.99, and the replay buffer size to 100,000. A mini-batch size of 64 was used to ensure stable learning. The detailed experimental settings for each algorithm are as follows:

\begin{itemize}

    \item QRL-NAS: Unlike existing QRL methods that use fixed quantum circuit structures based on researchers' intuition, QRL-NAS designs problem-optimized circuits by exploring various combinations of single-qubit and two-qubit gates. Gate types and arrangements are defined as search variables, and performance is evaluated through the RL process. This approach minimizes unnecessary gate usage while maximizing circuit expressiveness, improving both learning convergence speed and policy performance.
    

    \item QRL-DQN: QRL-DQN applies the DQN algorithm~\cite{DQN} with a fixed quantum circuit structure and performs value-based updates. A fixed PQC structure is used as the value function approximation model, with state information embedded via $RX$ gate-based angle encoding. Measurement outputs from PQC are interpreted as Q-values, and actions are selected using an $\epsilon$-greedy policy. Selected actions, rewards, and next states are used to update parameters via the DQN loss function.


    \item QRL-Reinforce: QRL-Reinforce applies the reinforce algorithm~\cite{williams1992simple} with a fixed quantum circuit structure and performs policy gradient-based updates. A fixed PQC structure is configured as the policy network, and state information is embedded using $RX$ gate-based angle encoding. Actions are sampled from the probability distribution obtained through measurement, and parameters are updated using the reinforce loss function based on collected rewards.

\end{itemize}

\subsection{Results of the Experiments}

\begin{figure}[t!]
    \begin{center}
        \includegraphics[width=3.3in]{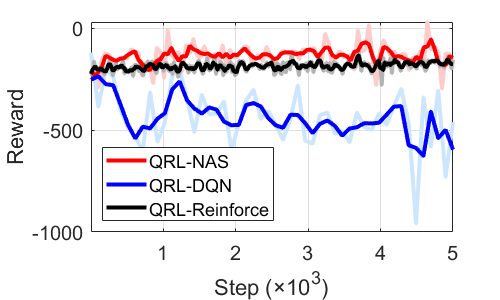}
    \end{center}
    \caption{Comparison between various QRL algorithms and the proposed QRL-NAS algorithm.}
    \label{fig:QRL-NAS_graph}
    \vspace{-3mm}
\end{figure}


Fig.~\ref{fig:QRL-NAS_graph} presents the reward changes at each learning steps for QRL-NAS, QRL-DQN, and QRL-Reinforce. QRL-NAS maintains higher rewards than the other algorithms from early learning stages, with average reward values gradually increasing over time. This indicates that QRL-NAS improves convergence speed and policy performance by exploring various gate combinations to optimize the quantum circuit structure for the task. In contrast, QRL-DQN shows large reward fluctuations, with average reward values remaining below -500, and fails to achieve stable convergence. This suggests that the fixed circuit structure used in QRL-DQN is unsuitable for the problem and limits learning efficiency. QRL-Reinforce shows low initial rewards but achieves stable convergence as rewards gradually increase during learning. However, its final average reward remains lower than that of QRL-NAS, indicating a limit to policy performance when using fixed circuits without structural exploration. In summary, QRL-NAS achieves faster convergence and higher rewards compared to QRL-DQN and QRL-Reinforce. The results demonstrate that introducing NAS to define and optimize gate types and arrangements effectively enhances QRL performance.

\section{Concluding Remarks} \label{sec:conclusion}

This paper proposes a QRL-NAS algorithm that integrates NAS to automatically optimize quantum circuit structures used as policy networks in QRL. Previous QRL papers primarily designed fixed PQC structures based on empirical intuition. In contrast, this paper defines a search space including various single-qubit and two-qubit gates and employs NAS to explore data-driven circuit structures optimized for problem environments. Experimental results on the LunarLander-v2 benchmark show that QRL-NAS achieves higher average rewards and faster convergence compared to existing QRL methods such as QRL-DQN and QRL-Reinforce. These results demonstrate that NAS-based quantum circuit structure optimization enhances both expressiveness and computational efficiency in QRL, maximizing learning stability and policy performance. Future research will focus on improving search efficiency by integrating one-shot NAS techniques such as ProxylessNAS and DARTS. Extensions to multi-environment and multi-task RL problems will also be explored. Additionally, feasibility evaluations on actual quantum hardware will be conducted. Performance assessments in noisy environments will also be carried out to validate QRL-NAS as a core technology for practical quantum artificial intelligence systems.

\bibliographystyle{IEEEtran}


\end{document}